\documentclass[journal]{IEEEtran}

\newenvironment{IEEEImpStatement}{\textbf{Implications Statement:}\begin{itshape}}{\end{itshape}}

\usepackage[colorlinks,urlcolor=blue,linkcolor=blue,citecolor=blue]{hyperref}

\usepackage{color,array}
\usepackage{cite}
\usepackage{amsmath,amssymb,amsfonts}
\usepackage{algorithm}
\usepackage{algorithmic}
\usepackage{graphicx}
\usepackage{textcomp}
\usepackage{xcolor}
\usepackage{tabularx}
\usepackage{multirow}
\usepackage{tikz}
\usetikzlibrary{positioning, shapes, backgrounds}
\usepackage{array}
\usepackage{rotating}
\usepackage{float}

\begin{document}

\title{Exploring Prompt Engineering: A Systematic Review with SWOT Analysis }

\author{Aditi Singh, Abul Ehtesham, Gaurav Kumar Gupta, Nikhil Kumar Chatta, Saket Kumar, Tala Talaei Khoei}

\markboth{}
{}

\maketitle

\begin{abstract}
In this paper, we conduct a comprehensive SWOT analysis of prompt engineering techniques within the realm of Large Language Models (LLMs). Emphasizing linguistic principles, we examine various techniques to identify their strengths, weaknesses, opportunities, and threats. Our findings provide insights into enhancing AI interactions and improving language model comprehension of human prompts. The analysis covers techniques including template-based approaches and fine-tuning, addressing the problems and challenges associated with each. The conclusion offers future research directions aimed at advancing the effectiveness of prompt engineering in optimizing human-machine communication.
\end{abstract}

\begin{IEEEImpStatement}
Prompt engineering enhances communication with Large Language Models (LLMs). Our SWOT analysis identifies the strengths, weaknesses, opportunities, and threats of various techniques, including template-based approaches and fine-tuning. By focusing on linguistic principles, we offer insights to improve AI interactions and comprehension of human prompts. This research advances AI capabilities and addresses challenges, paving the way for more effective human-AI communication. The findings benefit applications in customer service, education, and beyond, leading to more reliable and responsive AI systems.
\end{IEEEImpStatement}


\section{Introduction}

\IEEEPARstart{P}{rompt} engineering is a rapidly evolving field within artificial intelligence, particularly focused on optimizing the interaction between humans and Large Language Models (LLMs) \cite{marvin2023prompt, velasquez2023prompt, oppenlaender2023taxonomy}. At its core, prompt engineering involves designing and structuring inputs—known as prompts—to elicit the most accurate, relevant, and useful responses from AI systems. This practice is grounded in linguistic principles, leveraging an understanding of language patterns and structures to craft prompts that guide AI behavior effectively.
The emergence of Large Language Models \cite{singh2023exploring, brown2020,thoppilan2022lamdalanguagemodelsdialog,rae2022scalinglanguagemodelsmethods,chowdhery2022palmscalinglanguagemodeling} has highlighted the importance of prompt engineering \cite{gao2023prompt}. These models have demonstrated remarkable capabilities in generating human-like text, text-to-images, text-to-videos \cite{singh2023survey}, answering questions \cite{tan2023can, allemang2024increasing, hegde2023analyzing, arefeen2024leancontext,wang2023keqing}, and performing various language tasks \cite{van2024field,li2024chatcite, ramprasad2024analyzing, jin2024comprehensive}. However, their performance heavily depends on how well the prompts are crafted. Effective prompt engineering can significantly enhance the accuracy and relevance of AI responses, making the interaction more intuitive and productive.
Various techniques have been developed to refine prompt engineering, including template-based approaches, where fixed structures are used to standardize prompts, and fine-tuning methods that adapt the model to specific tasks or domains \cite{ye2023prompt, sun2023or, jin2024apeer}. These techniques aim to mitigate common issues such as ambiguity, bias, and context sensitivity, thereby improving the robustness and reliability of AI outputs. As AI continues to integrate more deeply into everyday applications, the role of prompt engineering becomes increasingly vital in ensuring seamless and meaningful human-AI communication. 
\subsection{Key Findings}
The key findings of the paper are as follows:
\begin{itemize}
\item \textit{Synergies:} Identified synergies between AI, Linguistics and Prompt Engineering.
\item \textit{Techniques:} Identified and categorized numerous prompt engineering methods.
\item \textit{Metrics:} Identified numerous metrics for evaluation of different prompt engineering methods, including BLEU, BERTScore, ROUGE, and Perplexity.
\item \textit{SWOT Analysis:} Identified strengths, weaknesses, opportunities, and threats for various prompt engineering techniques.
\end{itemize}

\section{Methodology}

This survey offers an extensive examination of the field of prompt engineering, incorporating insights from over 100 papers sourced from prominent academic databases and online platforms such as IEEE Xplore, ACM Digital Library, Google Scholar, and more. Queries utilizing keywords related to prompt engineering were employed to gather a comprehensive set of publications.

\section{Background}

\subsection{Prompt Engineering}

Prompt engineering involves crafting tailored instructions or prompts to direct the responses of advanced language models, such as GPT-3, towards a specific outcome (for instance, instructing ChatGPT to produce a particular text) \cite{Ali2023}. Prompt engineering involves designing input prompts that elicit accurate and valuable responses from large language models (LLMs) \cite{Petroni2020H}.
Prompt engineering refers to the practice of crafting and improving input queries, known as "prompts," to obtain specific outcomes from Large Language Models (LLMs). These prompts play a key role in directing LLMs to produce outputs that are both relevant and beneficial \cite{Oppenlaender2022}.
Prompt engineering creates a method for designing prompts that solve different problems, allowing for customization across various fields. It enhances LLM outputs by merging multiple prompt strategies and fosters knowledge sharing among users and developers of LLMs \cite{Lo2023}. 
Prompt engineering streamlines LLM application development, saving time and offering customizable interactions. It simplifies solving common issues, improving response accuracy and aiding conversational AI progress \cite{White2023}.
Prompt engineering is set to significantly enhance the capabilities of Large Language Models (LLMs), facilitating precise and swift language output. This emerging field not only promises to boost efficiency and optimize operations across sectors but also opens up new career paths for those proficient in prompt crafting. With ongoing advancements in sophisticated prompts, we can expect more intuitive user interfaces for LLM management, allowing for refined content generation and the exploration of previously unattainable LLM applications \cite{Abukhalaf2023, Oppenlaender2023}.
Prompt engineering enhances LLM applications by fostering a deeper comprehension of LLM behaviors and capabilities, guiding LLMs towards delivering truthful and informative responses. It boosts few-shot learning by integrating optimized prompts with traditional learning techniques, leading to more efficient chatbots, virtual assistants, and specialized prompt engineering tools for conversational AI. Thus, it plays a crucial role in advancing NLP tasks through improved LLM performance \cite{Short2023, Strobelt2023}.

Prompt engineering guides generative AI to desired outputs by crafting detailed instructions using specific words and formats. This creative process, involving trial and error, ensures AI interacts meaningfully with users and meets application expectations \cite{AWS2023}. 

\subsection{Linguistic Principles in Prompt Engineering}

Marjorie McShane and Sergei Nirenburg \cite{McShaneNirenburg2024} suggested that Linguistics for the age of AI rests on four major pillars: 
\begin{itemize}
\item Pillar 1: Development of language processing within a unified agent framework. 
\item Pillar 2: Human-inspired modeling for explanatory AI and actionable insights. 
\item Pillar 3: Contribution to and learning from linguistic scholarship. 
\item Pillar 4: Use of all heuristic evidence for meaning extraction and representation. 
\end{itemize}

The four pillars of Linguistics for AI proposed by Marjorie McShane and Sergei Nirenburg \cite{McShaneNirenburg2024} share similarities with several aspects of prompt engineering as defined in the descriptions provided:

\textit{Development within a unified agent framework (Pillar 1)} aligns with the goals of prompt engineering to streamline LLM application development and offer customizable interactions, improving the efficiency of language output and response accuracy \cite{White2023, Abukhalaf2023}. This connection highlights the integration of complex systems and the aim for coherence in both fields.

\textit{Human-inspired modeling for explanatory AI (Pillar 2)} mirrors the intention behind prompt engineering to foster a deeper comprehension of LLM behaviors and capabilities, guiding LLMs to deliver truthful and informative responses \cite{Short2023, Strobelt2023}. Both emphasize the importance of human-like understanding and reasoning in AI systems.

\textit{Learning from and contributing to linguistic scholarship (Pillar 3)} is paralleled by the aspect of prompt engineering that involves crafting and improving prompts based on trial and error, which necessitates an understanding of language and its nuances \cite{AWS2023}. This reflects a mutual interest in advancing linguistic knowledge and applying it to enhance AI capabilities.

\textit{Incorporation of heuristic evidence for meaning extraction (Pillar 4)} can be seen in the approach of prompt engineering to design prompts that elicit accurate and valuable responses from LLMs \cite{Petroni2020H}. Both areas utilize comprehensive data and insights to refine the interpretation and generation of language.

\begin{figure}[ht]
    \centering
    \includegraphics[width=0.50\textwidth]{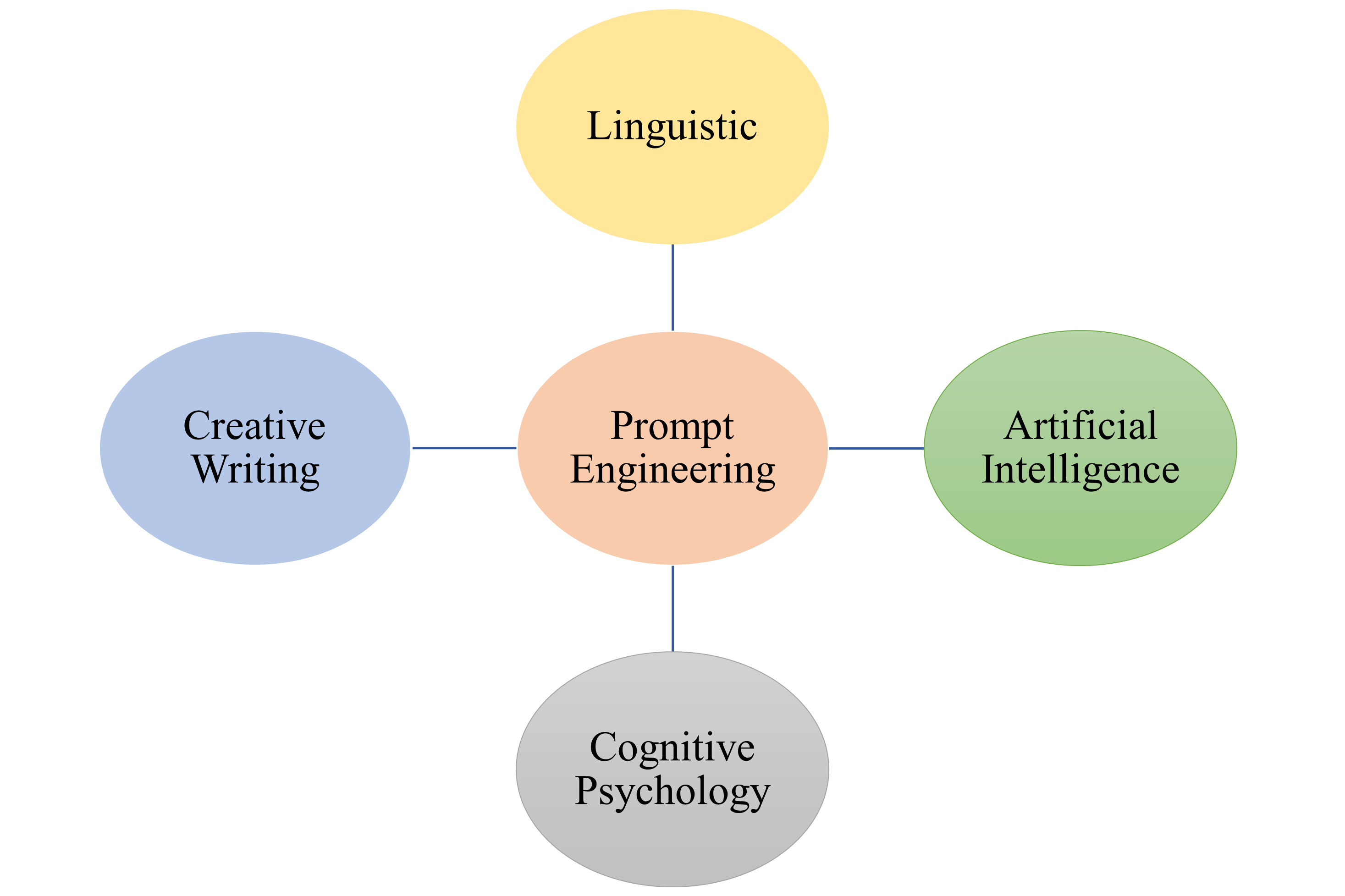}
    \caption{Convergence of AI, Linguistics, Psychology, and Creativity in Prompt Engineerings}
    \label{fig:ai_linguistic_pe}
\end{figure}

Fig \ref{fig:ai_linguistic_pe} and Table \ref{table:linguistics-ai-prompt-engineering} illustrate the convergence between AI linguistics and prompt engineering. 

\begin{table*}[ht!]
    \caption{Synergies between Linguistics for AI Pillars and Prompt Engineering Aspects}
    \label{table:linguistics-ai-prompt-engineering}
    \centering
    \fontsize{9}{11}\selectfont
    \renewcommand{\arraystretch}{1.2}  
    \begin{tabularx}{\textwidth}{|>{\raggedright\arraybackslash}p{0.28\textwidth}|X|}
    \hline
    \textbf{Linguistics for AI Pillar} & \textbf{Corresponding Aspect in Prompt Engineering} \\
    \hline
    Development of language processing within a unified agent framework. & Aligns with prompt engineering goals to streamline LLM development and enhance interaction efficiency, improving language output and accuracy. \\
    \hline
    Human-inspired modeling for explanatory AI and actionable insights. & Mirrors prompt engineering's aim to deepen understanding of LLM behaviors, guiding them to deliver truthful and informative responses. \\
    \hline
    Contribution to and learning from linguistic scholarship. & Informs prompt engineering by refining prompts through iterative testing, leveraging linguistic knowledge to enhance language nuance and accuracy. \\
    \hline
    Use of heuristic evidence for meaning extraction and representation. & Guides prompt engineering in crafting prompts that elicit accurate and valuable responses from LLMs, enhancing interpretation and generation of language. \\
    \hline
    \end{tabularx}
\end{table*}


\section{Related Works}

\subsection{Brief survey of prompt engineering techniques}
A number of surveys \cite{ Schulhoff2024ThePR, Sahoo2024, Chen2023, Gu2023} have been conducted to provide overviews and summaries of existing prompt engineering techniques, emphasizing advancements, applications, and practical insights (see Table \ref{table:prompt-engineering-survey}). However, our study differs by conducting a comprehensive Strength Weakness Opportunity and Threat (SWOT) analysis, focusing specifically on the strengths, weaknesses, opportunities, and threats associated with each technique. Additionally, we delve deeper into the linguistic principles shaping prompt design and offer targeted research directions to address current challenges and enhance future AI interactions.

\begin{table*}[ht]
    \caption{Summary of Surveys on Prompt Engineering Techniques}
    \label{table:prompt-engineering-survey}
    \centering
    \fontsize{8}{10}\selectfont
    \renewcommand\tabularxcolumn[1]{m{#1}} 
    \begin{tabularx}{\textwidth}{
      | >{\centering\arraybackslash}m{0.12\textwidth} 
      | >{\centering\arraybackslash}X
      | >{\centering\arraybackslash}X
      | >{\centering\arraybackslash}X | 
    }
    \hline
    Reference & 
    Highlights of Study & Strengths & Limitations \\
    \hline
    \cite{Schulhoff2024ThePR} & \begin{itemize}
        \item Establishes a comprehensive taxonomy and vocabulary for prompt engineering.
        \item Covers 58 text-based and 40 multimodal techniques.
        \item Facilitates rapid experimentation and standardization.
    \end{itemize} & 
    \begin{itemize}
        \item Comprehensive taxonomy and vocabulary standardization.
        \item Accessible and practical for rapid experimentation.
        \item Inclusive of multilingual and multimodal techniques.
    \end{itemize} & 
    \begin{itemize}
        \item Focuses on hard prompts, potentially overlooking benefits of soft prompts.
    \end{itemize} \\
    \hline
    \cite{Sahoo2024} & 
    \begin{itemize}
        \item Offers a detailed review of prompt engineering advancements.
        \item Summarizes methods, applications, models, and datasets.
        \item Includes taxonomy and tables for easy comparison.
    \end{itemize} & 
    \begin{itemize}
        \item Enhanced Model Efficacy: Demonstrates how prompt engineering can extend the capabilities of LLMs and VLMs without altering core model parameters.
        \item Application Diversity: Versatile application range.
        \item Foundation for future research.
    \end{itemize} & 
    \begin{itemize}
        \item While providing an overview, the study may not delve deeply into the technical details or theoretical underpinnings of each prompting method.
    \end{itemize} \\
    \hline
    \cite{Chen2023} & 
    \begin{itemize}
        \item Overview of prompt engineering for LLM optimization.
        \item Coverage of methodologies from basic to advanced.
        \item Role of external plugins in enhancing LLM performance.
    \end{itemize} & 
    \begin{itemize}
        \item Comprehensive introduction to prompt engineering.
        \item Insight into reducing LLM errors with plugins.
        \item Highlights future research avenues and practical applications.
    \end{itemize} & 
    \begin{itemize}
        \item Overlooks domain-specific challenges.
    \end{itemize} \\
    \hline
    \cite{Gu2023} &
    \begin{itemize}
        \item Comprehensive survey on prompt engineering in vision-language models (VLMs).
        \item Classifies prompting methods into hard and soft prompts.
        \item Discusses applications and ethical considerations.
    \end{itemize} & 
    \begin{itemize}
        \item Enables adaptation of large pre-trained models to new tasks with minimal data.
        \item Facilitates predictions based solely on prompts, preserving model parameters.
        \item Applicable across multimodal domains: text, image, and their combinations.
    \end{itemize} & 
    \begin{itemize}
        \item Dependency on high-quality prompts for effective task adaptation.
        \item Challenges in maintaining interpretability and ethical use of prompted models.
        \item Limited systematic overview of prompting methods across all VLM types.
    \end{itemize} \\
    \hline
    \end{tabularx}
\end{table*}

\subsection{Different types of Prompt Engineering Techniques}
The different types of prompt engineering techniques are as follows. A summary of each of these is provided in Table \ref{table:prompt-engineering-techniques}. 
\subsubsection{Automatic Reasoning and Tool-use}
Automatic Reasoning and Tool-use (ART) \cite{Sahoo2024, paranjape2023art} is a computational framework designed to augment the capabilities of Large Language Models (LLMs) for complex problem-solving in few-shot and zero-shot settings. ART combines LLM-generated "chain of thought" (CoT) reasoning with the execution of external tools, thereby enabling tasks that surpass standard linguistic processing. This integration allows ART to automate the generation of intermediate reasoning steps, formatted as executable programs, which strategically incorporate external data through tool interactions.

ART operates by selecting appropriate multi-step reasoning templates from a task library and dynamically incorporating responses from external tools into the LLM's workflow. This process is mathematically managed by pausing and resuming the LLM's output generation based on tool interaction points, formalized as:
\begin{equation}
\text{Output}{\text{final}} = f(\text{Output}{\text{LLM}}, \text{Tool}_{\text{output}})
\end{equation}
where $f$ denotes the function that integrates tool outputs into the LLM's reasoning process.

\subsubsection{Chain-of-Thought (CoT)}
Chain-of-Thought (CoT) Prompting \cite{wei2022chain, Sahoo2024, Chen2023, zhang2023automatic, kojima2023largelanguagemodelszeroshot} is a technique designed to facilitate complex reasoning by generating intermediate reasoning steps. This approach allows Large Language Models (LLMs) to articulate their thought processes step-by-step, thereby enhancing their ability to tackle more intricate tasks that demand preliminary reasoning before producing a response. The Chain-of-Thought Prompting can be represented as:

\begin{equation}
\begin{aligned}
K_0 &= s, \\
r_1 &= g(s, K_0), \\
r_2 &= g(s, K_1), \\
&\vdots \\
r_n &= g(s, K_{n-1}), \\
K_i &= K_{i-1} \cup \{r_i\}, \\
\text{Output}_{\text{final}} &= h(K_n),
\end{aligned}
\end{equation}
Here, 
\begin{itemize}
  \item $K_0$ is defined as the starting point containing the initial problem statement $s$.
  \item $r_i$ denotes each reasoning step, where $g$ is a function modeling the LLM's processing to generate the reasoning step based on the current state of knowledge $K_{i-1}$ and the initial problem statement $s$.
   \item $K_i$ accumulates each reasoning step into the knowledge base, effectively building upon each previous step.
  \item $\text{Output}_{\text{final}}$ represents the final outcome, which $h$ computes from the fully accumulated knowledge $K_n$ after all reasoning steps.
\end{itemize}

CoT prompting has been adapted for use in multilingual contexts through several innovative approaches. One such method is XLT (Cross-Lingual Thought) Prompting, developed by Huang et al. \cite{Huang2023NotAL}, which utilizes a prompt template incorporating six distinct instructions, including role assignment, cross-lingual reasoning, and CoT. Additionally, Cross-Lingual Self Consistent Prompting (CLSP), proposed by Qin et al. \cite{Qin2023CrosslingualPI} (2023a), employs an ensemble technique to construct reasoning paths in diverse languages, further broadening the applicability and effectiveness of CoT prompting in multilingual settings.
While Chain-of-Thought (CoT) prompting has shown substantial success in English, its application in low-resource languages remains limited. To address this gap, Chai et al. developed xCoT, a framework that transfers knowledge from high-resource to low-resource languages, enhancing multilingual CoT reasoning capabilities
\cite{Chai2024xCoTCI}.
Despite progress in cross-lingual Chain-of-Thought (CoT) reasoning, existing methods face limitations due to the manual specification of languages and static weight allocation across different language reasoning paths. To overcome these challenges, Zhang introduces AutoCAP, a framework that automates language selection and dynamically allocates weight scores to different reasoning paths for zero-shot CoT, significantly enhancing performance and generalizability
\cite{Zhang2024AutoCAPTA}. Shi et al. \cite{Shi2022LanguageMA} explore the multilingual reasoning abilities of large language models through Chain-of-Thought (CoT) prompting, demonstrating that their effectiveness in solving diverse language tasks, such as the Multilingual Grade School Math (MGSM) benchmark, enhances with model scale and extends to both well-represented and underrepresented languages. However, the study highlights a critical gap in the dependency on model size for robust multilingual performance using CoT prompting, underscoring the need for more efficient architectures or training strategies that can achieve similar results without extensive scaling. Chen et al. \cite{Chen2024M3CoTAN} address critical gaps in existing multi-modal Chain-of-Thought (CoT) benchmarks by introducing a novel benchmark that incorporates multi-domain, multi-step, and multi-modal reasoning capabilities. Despite these advancements, their findings reveal that Vision Large Language Models (VLLMs) struggle to perform accurately within this complex CoT framework, highlighting a significant performance disparity between VLLMs and human capabilities. This pioneering work sets a foundation for future exploration and enhancement of multi-modal reasoning systems.

\subsubsection{Directional Stimulus Prompting}
Directional Stimulus Prompting (DSP) \cite{NEURIPS2023_c5601d99} is a method in prompt engineering that embeds specific guidance or stimuli within prompts to direct the language model's responses towards a desired outcome. This method enhances the model's performance and relevance by including subtle cues or explicit instructions alongside the task description.

In DSP, discrete tokens known as "directional stimuli" are introduced into the prompt to guide the model. For instance, in a summarization task, these stimuli might include essential keywords to be reflected in the summary.

The mathematical representation of this process is as follows:
\begin{equation}
y \sim p_{LLM}(y \mid x, z),
\end{equation}
\begin{equation}
z \sim p_{POL}(z \mid x).
\end{equation}

Where:
\begin{itemize}
  \item \( x \) is the original input.
  \item \( p_{POL}(z \mid x) \) is a policy language model generating the directional stimulus \( z \) from \( x \).
  \item \( p_{LLM}(y \mid x, z) \) is the language model generating the output \( y \) based on both the input \( x \) and the directional stimulus \( z \).
\end{itemize}

The parameters of \( p_{LLM} \) remain unchanged, maintaining the efficiency and stability of the model while providing precise guidance through the additional stimuli.

\subsubsection{Few-Shot Prompting}
Few-shot prompting \cite{Sahoo2024, Chen2023,xu2023making, dhamani-engler-2024, Lee2024, Timo2022, rubin-etal-2022-learning, su2022selective} is a technique that enhances in-context learning by including example demonstrations within the prompt. These examples guide the model to generate accurate responses for subsequent tasks based on the provided context.
In few-shot prompting, the model's behavior is influenced by a small number of examples. Let’s denote:

\begin{itemize}
  \item \( x \) as the original input or query.
  \item \( \{(x_i, y_i)\}_{i=1}^k \) as the set of \( k \) few-shot examples, where each example consists of an input \( x_i \) and a corresponding output \( y_i \).
  \item \( p_{LM}(y \mid x, \{(x_i, y_i)\}_{i=1}^k) \) as the language model generating the output \( y \) based on the input \( x \) and the few-shot examples.
\end{itemize}

The mathematical representation of few-shot prompting can be written as:
\begin{equation}
y \sim p_{LM}(y \mid x, \{(x_i, y_i)\}_{i=1}^k),
\end{equation}
where \( \{(x_i, y_i)\}_{i=1}^k \) are the \( k \) few-shot examples provided to the model to guide its generation of the output \( y \).

Lee et al. \cite{Lee2024} explores the efficacy of ChatGPT and prompt engineering for automatic question generation in English education, demonstrating significant improvements in question validity through few-shot prompting techniques. However, it highlights a gap in the optimization of certain question types via few-shot prompting, indicating a need for further refinement to enhance the versatility and reliability of AI-generated educational content.
Timo Schick and Hinrich Schütze \cite{Timo2022} demonstrate that the Pet method, which combines textual instructions with example-based finetuning, performs strongly in true few-shot settings without requiring a development set, achieving new state-of-the-art results on the RAFT benchmark. However, the study highlights a gap in understanding the specific design choices and configurations necessary for optimal performance in true few-shot learning scenarios, indicating a need for further research into intelligent prompt handling and configuration.
Xi Ye and Greg Durrett \cite{NEURIPS2022_c4025018} investigated whether prompting large language models (LLMs) like GPT-3 with explanations enhances in-context learning for textual reasoning tasks. Their study finds that while explanations provide small to moderate accuracy improvements for most models, text-davinci-002 benefits more significantly. However, explanations often lack alignment with the models' predictions or factual grounding. 
Chengyu Wang et. al. \cite{wang-etal-2021-transprompt} introduced TransPrompt, a framework that leverages transferable prompt embeddings for few-shot text classification across similar NLP tasks. TransPrompt uses a meta-learner trained through a multi-task meta-knowledge acquisition process, employing de-biasing techniques to remain task-agnostic. Extensive experiments show TransPrompt outperforms strong baselines. However, optimizing transferability and de-biasing techniques for varied tasks remains a challenge, requiring further research.
\subsubsection{Generated Knowledge Prompting}

Generated knowledge prompting \cite{liu2021generated} is a technique that involves enhancing a language model's performance on a multiple-choice commonsense reasoning task through two key steps: Knowledge Generation and Knowledge Integration.

In such tasks, we predict an answer \( \hat{a} \in A_q \) given a question \( q \in Q \), where \( A_q \) is the set of choices for the question \( q \). The method involves two steps:

1. Knowledge Generation:
   Generate knowledge statements \( K_q \) conditioned on the question:
   \begin{equation}
   K_q = \{k_m \mid k_m \sim p_G(k \mid q), \; m = 1, \ldots, M\}.
   \end{equation}
   \begin{itemize}
     \item \( K_q \): The set of generated knowledge statements related to the question \( q \).
     \item \( k_m \): An individual knowledge statement generated for the question \( q \).
     \item \( p_G(k \mid q) \): The probability distribution used to generate the knowledge statements \( k_m \) given the question \( q \).
     \item \( m \): An index denoting the different knowledge statements in the set \( K_q \).
     \item \( M \): The total number of knowledge statements generated.
   \end{itemize}

2. Knowledge Integration:
   Integrate the generated knowledge into the decision process for inference:
   \begin{equation}
   \hat{a} = \arg \max_{a \in A_q} p_I(a \mid q, K_q).
   \end{equation}
   \begin{itemize}
     \item \( \hat{a} \): The predicted answer to the question \( q \).
     \item \( A_q \): The set of possible answer choices for the question \( q \).
     \item \( p_I(a \mid q, K_q) \): The probability of answer \( a \) given the question \( q \) and the generated knowledge \( K_q \).
     \item \( \arg \max_{a \in A_q} \): The operation to find the answer \( a \) that maximizes the probability.
   \end{itemize}

In comparison, without using generated knowledge, the inference model yields:
\begin{equation}
\hat{a} = \arg \max_{a \in A_q} p_I(a \mid q).
\end{equation}
 \begin{itemize}
     \item \( \hat{a} \): The predicted answer to the question \( q \) without additional generated knowledge.
     \item \( p_I(a \mid q) \): The probability of answer \( a \) given only the question \( q \), without the generated knowledge \( K_q \).
     \item \( \arg \max_{a \in A_q} \): The operation to find the answer \( a \) that maximizes the probability based solely on the question \( q \).
   \end{itemize}
Building on the concept of knowledge prompting, Jiajin Tang et al. \cite{Tang_2023_ICCV} introduced a framework called CoTDet, which integrates knowledge prompting with chain-of-thought reasoning for task-driven object detection. CoTDet uses knowledge prompting to extract and apply essential affordance knowledge from large language models, focusing on attributes that enable various objects to perform specific tasks. It then employs multi-level chain-of-thought (MLCoT) reasoning to systematically link this knowledge to object attributes through rationales. This combination enhances object detection and localization, with CoTDet achieving significant improvements in both box and mask AP compared to existing methods.

Similarly, Jianing Wang et al. \cite{Wang2022KnowledgePI} introduced KP-PLM, a framework for enhancing pre-trained language models with factual knowledge using natural language prompts. This approach avoids complex modifications to PLM architectures and redundant information from knowledge bases. KP-PLM employs a knowledge sub-graph and two self-supervised tasks to improve performance. Experiments show it outperforms current methods in natural language understanding tasks.

Additionally, Jianing Wang et al. \cite{Wang2023BoostingLM} proposed Chain-of-Knowledge (CoK) prompting to address the limitations of Chain-of-Thought (CoT) prompting in reasoning tasks. CoK prompting aims to elicit explicit knowledge evidence in the form of structured triples, inspired by human reasoning processes. To enhance reliability, the authors introduce the F²-Verification method to assess the factuality and faithfulness of reasoning chains, prompting the model to reconsider unreliable responses. Extensive experiments show that CoK prompting further boosts performance across various reasoning tasks, including commonsense, factual, symbolic, and arithmetic reasoning. Lihui Zhang and Ruifan Li \cite{10.1007/978-981-99-8145-8_3} presented the Knowledge Prompting with Contrastive Learning (KPCL) model for unsupervised commonsense question answering. KPCL improves performance by using dropout noise for augmentation, unsupervised contrastive learning for nuanced question handling, and generic prompts for zero-shot knowledge generation.
Xiaohan Zhang et al. \cite{Zhang2024DKPROMPTDK} combined knowledge prompting with vision-language models through their framework, DKPROMPT. This method integrates domain knowledge from PDDL-based classical planning to enhance VLMs for open-world task planning. DKPROMPT effectively bridges the gap between VLMs' vision-language capabilities and classical planning's robustness, leading to superior task completion rates compared to traditional and VLM-only methods.

\subsubsection{Graph Prompting}

Graph Prompting \cite{Wu2023ASO, Yi2023, tian2024graph, sun2023graph, sun2022, xu2024gipcol,jing2023deep,sunchengli2023} is a technique in prompt engineering that leverages graph data to create more effective prompts for machine learning models. It reformulates tasks to resemble pretext tasks, enabling the direct use of pre-trained models. This method integrates relational and contextual information from graphs, enhancing the prompts' precision and relevance. There are two primary types of Graph Prompting:
\begin{itemize}
    \item \textit{Discrete Prompts:} These utilize natural language or specific graph elements to create prompts. They involve manually or automatically crafted templates that incorporate graph-based knowledge, making them suitable for tasks that require explicit contextual information.
    \item \textit{Continuous Prompts:} These involve learned representations or embeddings. Continuous prompts dynamically adjust the input data in the embedding space, utilizing graph representation learning methods to generate contextually enriched prompts.
\end{itemize}

The mathematical framework for Graph Prompting can be outlined as follows:

\begin{equation}
x' = f_{\text{prompt}}(x; \theta_{\text{prompt}})
\end{equation}
Where \( x \) is the input sample, and \( \theta_{\text{prompt}} \) represents the task-related knowledge parameters incorporated into the prompt.

For the pre-training phase, a link prediction task to learn generalizable knowledge from a graph \( G = (V, E) \) is employed:
\begin{equation}
\begin{aligned}
V(S_v) &= \{ u \in V \mid d(u, v) \leq \delta \} \\
E(S_v) &= \{ (u, u') \in E \mid u \in V(S_v), u' \in V(S_v) \}
\end{aligned}
\end{equation}
Where \( S_v \) is the contextual subgraph of node \( v \), and \( \delta \) is a predetermined threshold.

The subgraph representation \( s_x \) is computed using a ReadOut operation:
\begin{equation}
s_x = \text{ReadOut}(\{ h_v : v \in V(S_x) \})
\end{equation}

For downstream tasks, such as link prediction, node classification, and graph classification, the following formulations is used:
\begin{equation}
\text{sim}(s_v, s_a) > \text{sim}(s_v, s_b)
\end{equation}
\begin{equation}
\ell_j = \arg\max_{c \in C} \text{sim}(s_{v_j}, \tilde{s}_c)
\end{equation}
\begin{equation}
L_j = \arg\max_{c \in C} \text{sim}(s_{G_j}, \tilde{s}_c)
\end{equation}

Learnable prompts can further refine the subgraph representations for specific tasks:
\begin{equation}
s_{t,x} = \text{ReadOut}(\{ p_t \odot h_v : v \in V(S_x) \})
\end{equation}
Where \( p_t \) is a learnable prompt vector, and \( \odot \) denotes element-wise multiplication.

Finally, the most probable answer for a given prompt is determined through:
\begin{equation}
z = \arg\max_{z' \in Z} P(f(x'), z')
\end{equation}
Where \( Z \) is the set of possible answers, and \( P \) is the probability or similarity function.

Graph Prompting effectively utilizes graph structures to generate contextually enriched prompts, improving the performance and adaptability of machine learning models in various graph-related tasks.

\subsubsection{Iterative Prompting}

Iterative prompting \cite{Liang_2023_ICCV, wang2022iteratively,krishna2024understanding,nasiriany2024pivotiterativevisualprompting,Chen2023IterativePR} is a method in prompt engineering where the prompts given to a Generative AI tool are progressively refined to enhance the relevance, accuracy, and depth of its responses. This approach is similar to a conversational exchange, where each answer helps shape the next question, allowing for continuous learning and adjustment based on feedback.

In iterative prompting, the process involves several essential steps:
\begin{enumerate}
    \item \textit{Initial Prompt:} Begin with a broad, open-ended prompt to assess the AI's initial understanding of the task.
    \item \textit{Response Analysis:} Examine the AI's responses for relevance and depth, identifying gaps or areas that need improvement.
    \item \textit{Prompt Refinement:} Adjust the prompt based on initial responses, incorporating specific keywords or phrases that were particularly insightful or relevant.
    \item \textit{Feedback Loop:} Treat the process as a continuous feedback loop, where each iteration of prompting is informed by the responses from previous iterations.
    \item \textit{Experimental Testing:} Test different prompt styles and validate the refined prompts on multiple examples to ensure robustness and effectiveness.
\end{enumerate}

The iterative prompting process can be mathematically represented as follows:
Given $x_0$ (initial input), Initialize:
\begin{equation}
    p_0 = f_{\text{prompt}}(x_0),
\end{equation}

\textit{For each iteration } t \text{ do:}
\begin{equation}
    \begin{aligned}
        &r_t = f_{\text{response}}(p_t), \\
        &e_t = f_{\text{error}}(r_t), \\
        &p_{t+1} = f_{\text{refine}}(p_t, e_t),
    \end{aligned}
\end{equation}

\textit{Until convergence or maximum iterations.}

\textit{Output:}
\begin{equation}
    y = f_{\text{final}}(r_T),
\end{equation}

Where:
\begin{itemize}
    \item $x_0$: Initial input or problem statement.
    \item $p_t$: Prompt at iteration $t$.
    \item $r_t$: Response from the AI model at iteration $t$.
    \item $e_t$: Error or feedback at iteration $t$.
    \item $f_{\text{prompt}}$: Function generating the initial prompt.
    \item $f_{\text{response}}$: Function generating the AI's response.
    \item $f_{\text{error}}$: Function evaluating the response to identify errors.
    \item $f_{\text{refine}}$: Function refining the prompt based on errors.
    \item $f_{\text{final}}$: Function producing the final output.
\end{itemize}

Iterative prompting, akin to iterative research, focuses on continuous improvement through design, learning, and refinement, ensuring the AI tool aligns accurately with research objectives and enhances the efficiency and effectiveness of data analysis.

\subsubsection{Least-To-Most Prompting}

Least-to-most prompting \cite{libby2008comparison, yanardaug2011effects, foran2021comparison, zhou2022least,murzynski2007combining,aljehany2020comparison,cheng2024mostbuildingplugandplayvisual} is a technique in prompt engineering that teaches language models to solve complex problems by breaking them down into simpler subproblems. It involves two main stages: 
\textit{Decomposition}: The initial prompt demonstrates how to decompose a complex problem into manageable subproblems.

\textit{Subproblem Solving}: The subsequent prompts guide the model to solve each subproblem sequentially until the original problem is solved.

This method can be mathematically represented as follows:

\textit{Stage 1: Decomposition:}
\begin{equation}
    D = \{d_1, d_2, \ldots, d_n\} = f_{\text{decompose}}(P),
\end{equation}
where $D$ is the set of subproblems $\{d_1, d_2, \ldots, d_n\}$, and $f_{\text{decompose}}$ is the decomposition function applied to the original problem $P$.

\textit{Stage 2: Subproblem Solving:}
\begin{equation}
    s_i = f_{\text{solve}}(d_i | \{s_1, s_2, \ldots, s_{i-1}\}), \text{ for } i = 1, 2, \ldots, n,
\end{equation}
where $s_i$ is the solution to subproblem $d_i$, and $f_{\text{solve}}$ is the solving function considering previous solutions.

The overall process ensures that the language model can handle complex problems by systematically addressing each subproblem. This can be summarized as:

\begin{equation}
\begin{aligned}
    \text{Final Solution: } S = \{s_1, s_2, \ldots, s_n\},
\end{aligned}
\end{equation}

where \(S\) is the set of solutions to all subproblems, providing the final answer to the original problem.

Least-to-most prompting can be combined with other techniques such as chain-of-thought and self-consistency, enhancing its effectiveness.

\subsubsection{Multimodal CoT prompting}

Multimodal Chain-of-Thought (CoT) Prompting \cite{Chen2024M3CoTAN, mitra2024compositional,mondal2024kam,lu2022learn,zheng2023ddcot,anand2024,shao2024visual,hu2024multimodal,gao2024cantor} is a technique that enhances language models' reasoning by integrating textual and visual inputs. It iteratively guides the model through reasoning steps that combine information from multiple modalities, leading to more comprehensive and contextually accurate responses.
The following mathematical representation includes CoT with Visual input:

\textit{Step 1: Initial Integration:} Given text input $T$ and visual input $V$,
\begin{equation}
    I_0 = f_{\text{integrate}}(T, V),
\end{equation}
where $I_0$ is the initial combined representation.

\textit{Step 2: Iterative Reasoning:} For $i = 1, 2, \ldots, n$,
\begin{equation}
    r_i = f_{\text{reason}}(I_{i-1}, \{r_1, r_2, \ldots, r_{i-1}\}),
\end{equation}
where $r_i$ is the reasoning step based on the previous state and prior steps, and $f_{\text{reason}}$ is the reasoning function.

\begin{equation}
    I_i = f_{\text{update}}(I_{i-1}, r_i),
\end{equation}
where $I_i$ is the updated combined representation.

\textit{Step 3: Final Output:}
\begin{equation}
    O = f_{\text{output}}(I_n),
\end{equation}
where $O$ is the final output from the final representation $I_n$.

\textit{Step 1: Initial Integration:} Given text input $T$ and visual input $V$,
\begin{equation}
    I_0 = f_{\text{integrate}}(T, V),
\end{equation}
where $I_0$ is the initial combined representation.

\textit{Step 2: Iterative Reasoning:} For $i = 1, 2, \ldots, n$,
\begin{equation}
    r_i = f_{\text{reason}}(I_{i-1}, \{r_1, r_2, \ldots, r_{i-1}\}),
\end{equation}
where $r_i$ is the reasoning step based on the previous state and prior steps, and $f_{\text{reason}}$ is the reasoning function.

\begin{equation}
    I_i = f_{\text{update}}(I_{i-1}, r_i),
\end{equation}
where $I_i$ is the updated combined representation.

\textit{Step 3: Final Output:}
\begin{equation}
    O = f_{\text{output}}(I_n),
\end{equation}
where $O$ is the final output from the final representation $I_n$.

In this representation:
\begin{itemize}
    \item \(T\) is the text input.
    \item \(V\) is the visual input.
    \item \(I_0\) is the initial integrated representation.
    \item \(r_i\) represents each reasoning step.
    \item \(I_i\) is the integrated representation after \(i\) iterations.
    \item \(O\) is the final output.
    \item Functions \(f_{\text{integrate}}\), \(f_{\text{reason}}\), and \(f_{\text{update}}\) handle integration, reasoning, and updating processes.
\end{itemize}

\subsubsection{ReAct}

ReAct Prompting \cite{yao2022react, verma2024brittle} ensures language models generate both reasoning traces and task-specific actions in an interleaved manner. This approach allows models to dynamically adjust their action plans and interact with external sources for additional information, enhancing reliability and factual accuracy. The formulation of ReAct Prompting is represented as:

\textit{Stage 1: Reasoning Trace Generation:} Given a task $T$ and initial prompt $P$,
\begin{equation}
    r_i = f_{\text{reason}}(P, \{a_1, \ldots, a_{i-1}\}),
\end{equation}
where $r_i$ is the reasoning step and $a_j$ are previous actions.

\textit{Stage 2: Action Execution:}
\begin{equation}
    a_i = f_{\text{act}}(r_i),
\end{equation}
where $a_i$ is the action derived from $r_i$.

\textit{Stage 3: Observation Integration:}
\begin{equation}
    o_i = f_{\text{observe}}(a_i),
\end{equation}
where $o_i$ is the observation from $a_i$.

\textit{Stage 4: Update Prompt:}
\begin{equation}
    P = P \cup \{r_i, a_i, o_i\},
\end{equation}
updating the prompt with the latest steps.

Repeat stages 1-4 until task completion or maximum iterations.

In this representation:

\begin{itemize}
    \item \(T\) is the given task.
    \item \(P\) is the initial prompt.
    \item \(r_i\) represents the \(i\)-th reasoning step generated by the reasoning function \(f_{\text{reason}}\).
    \item \(a_i\) represents the \(i\)-th action generated by the action function \(f_{\text{act}}\).
    \item \(o_i\) represents the \(i\)-th observation generated by the observation function \(f_{\text{observe}}\).
    \item The prompt \(P\) is iteratively updated with each new reasoning step, action, and observation.
\end{itemize}

\subsubsection{Self-Ask Prompting}

Self-Ask Prompting \cite{press2022measuring, Li2022SelfPromptingLL} is an advanced technique derived from Chain Of Thought (CoT) Prompting. It enhances the ability of language models (LMs) to answer complex questions by generating and answering sub-questions before addressing the main question. This method improves performance by breaking down a problem into manageable parts and systematically solving each part.

\textit{Stage 1: Decomposition:} Given a task $T$ and initial prompt $P$, 
\begin{equation}
    D = \{d_1, d_2, \ldots, d_n\} = f_{\text{decompose}}(P),
\end{equation}
where $D$ is the set of subproblems $\{d_1, d_2, \ldots, d_n\}$, and $f_{\text{decompose}}$ is the decomposition function applied to the original problem $P$.

\textit{Stage 2: Subproblem Solving}
\begin{equation}
    s_i = f_{\text{solve}}(d_i | \{s_1, s_2, \ldots, s_{i-1}\})
\end{equation}
for $i = 1, 2, \ldots, n, $ where $s_i$ is the solution to subproblem $d_i$, and $f_solve$ is the solving function considering previous solutions.

\textit{Stage 3: Integration:}
\begin{equation}
    \text{Final Answer} = f_{\text{integrate}}(\{s_1, s_2, \ldots, s_n\}),
\end{equation}
where $f_{\text{integrate}}$ is the integration function combining all subproblem solutions.

\subsubsection{Self-Consistency}

Self-consistency \cite{Sahoo2024, Chen2023,dhamani-engler-2024} is a technique in prompt engineering that improves the accuracy of language models by generating multiple candidate outputs for a given prompt and aggregating the results. This approach leverages diverse reasoning paths to enhance answer reliability.

Given a prompt and a question, self-consistency introduces a latent variable $r_i$, where $r_i$ represents the reasoning path in the $i$-th output, leading to the answer $a_i$. The final answer is chosen based on majority voting over the candidate answers $a_1, a_2, \ldots, a_m$.

\begin{equation}
    \text{Final answer} = \arg \max_{a} \sum_{i=1}^{m} \mathbf{1}(a_i = a),
\end{equation}
where $\mathbf{1}(a_i = a)$ is 1 if $a_i = a$ and 0 otherwise.

In more detail, assume the generated answers $a_i$ are from a fixed answer set, $a_i \in A$, where $i = 1, \ldots, m$ indexes the $m$ candidate outputs sampled from the decoder. Given a prompt and a question, self-consistency introduces an additional latent variable $r_i$, which is a sequence of tokens representing the reasoning path in the $i$-th output. This couples the generation of $(r_i, a_i)$ where $r_i \rightarrow a_i$, i.e., generating a reasoning path $r_i$ is optional and only used to reach the final answer $a_i$.

The method can be formally expressed as:

\begin{equation}
\begin{aligned}
    P(r_i, a_i \mid \text{prompt, question}) 
    &= P(a_i \mid r_i, \text{prompt, question}) \\
    &\quad \cdot P(r_i \mid \text{prompt, question}), \\
    \text{for } i &= 1, 2, \ldots, n.
\end{aligned}
\end{equation}

where the joint probability $P(r_i, a_i | \text{prompt, question})$ is decomposed into the probability of generating the answer given the reasoning path and the prompt, and the probability of the reasoning path given the prompt and question.

To compute $P(r_i, a_i | \text{prompt, question})$, we can either take the unnormalized probability of the model generating $(r_i, a_i)$ given (prompt, question), or we can normalize the conditional probability by the output length:


\begin{equation}
\begin{aligned}
    P(r_i, a_i | \text{prompt, question}) &=  \\
    \exp \left( \frac{1}{K} \sum_{k=1}^{K} \log P (t_k | \text{prompt, question}, t_1, \ldots, t_{k-1}) \right)
\end{aligned}
\end{equation}

where $\log P(t_k | \text{prompt, question}, t_1, \ldots, t_{k-1})$ is the log probability of generating the $k$-th token $t_k$ in $(r_i, a_i)$ conditioned on the previous tokens, and $K$ is the total number of tokens in $(r_i, a_i)$.

Self-consistency can be applied to problems where the final answer is from a fixed answer set. By introducing diversity in the reasoning processes, this technique enhances the robustness and accuracy of the language models' outputs.

\subsubsection{Sequential Prompting}

Sequential prompting \cite{Liu2024, Wu2024, lee2024improve} is a strategy used in natural language processing tasks to improve the accuracy of predictions by using the results of previous steps as prior knowledge for the next prediction.

In this approach, the task involves extracting elements \(e_i\) based on previous predictions and the initial input. The process uses the output of one step as an input prompt for the next step.

Given an input \(X = [x_1, x_2, \ldots, x_m]\), the goal is to extract a collection of elements \(E = \{e_i\}_{i=1}^{|E|}\).

\begin{itemize}
    \item \textit{Initial Extraction}:
    \begin{equation}
        e_1 = \text{argmax}_{e} \ P(e|X)
    \end{equation}

    \item \textit{Subsequent Predictions}:
    \begin{equation}
        e_i = \text{argmax}_{e} \ P(e|e_1, e_2, \ldots, e_{i-1}, X)
    \end{equation}
\end{itemize}

Here, \(P(e|X)\) is the probability of element \(e\) given the input \(X\), and \(P(e|e_1, e_2, \ldots, e_{i-1}, X)\) is the probability of element \(e\) given the previous elements \((e_1, e_2, \ldots, e_{i-1})\) and the input \(X\).

The sequential prompting strategy leverages these conditional probabilities to iteratively refine the predictions using the prior results.

\begin{table*}[ht!]
    \caption{Summary of Various Prompt Engineering Techniques}
    \label{table:prompt-engineering-techniques}
    \centering
    \fontsize{8}{10}\selectfont
    \renewcommand\tabularxcolumn[1]{m{#1}} 
    \begin{tabularx}{\textwidth}{
      | >{\centering\arraybackslash}m{0.1\textwidth} 
      | >{\centering\arraybackslash}X
      | >{\centering\arraybackslash}X
      | >{\centering\arraybackslash}X
      | >{\centering\arraybackslash}X | 
    }
    \hline
    Technique Name & Strength & Weakness & Opportunity & Threat \\
    \hline
    Automatic Reasoning and Tool-use (ART) & 
    Combines reasoning with tool-use; Addresses wide problems. & 
    Dependency on tools; Integration challenges. & 
    Advances AI autonomy; Expands real-world application. & 
    Reliability of tools; Integration challenges with systems. \\
    \hline
    Chain-of-Thought (CoT) & 
    Simplifies complex problems; Improves interpretability. & 
    Careful crafting needed; Possible inaccuracies. & 
    Enhances AI problem-solving. & 
    Resource-intensive; Overreliance risk. \\
    \hline
    Directional Stimulus Prompting & 
    Provides fine-grained guidance using discrete tokens to steer LLMs effectively. Enhances control and interpretability of LLM behaviors without modifying parameters. & 
    Depends on policy model quality, implementation complexity, and heuristic stimulus selection. Faces challenges in domain-specific adaptation. & 
    Optimization potential through reinforcement learning. Expands beyond text tasks, enhancing task-specific performance. & 
    Varied performance across tasks and LLM configurations. Ethical concerns about bias and fairness. Competes with other prompt techniques. \\

    \hline
    Few-Shot Prompting & 
    Minimal examples needed; CP-Tuning innovation. & 
    Complex prompt crafting; Bias risk. & 
    Streamlines few-shot learning. & 
    May be outpaced by LLM advancements. \\
    \hline
    
    Generated Knowledge Prompting &
    Improves task performance on commonsense benchmarks without structured knowledge bases or joint fine-tuning. &
    Performance variability based on knowledge quality and model capability. &
    Potential for broader application across tasks, simplifying adaptation without extensive retraining. &
    Faces competition from retrieval-based systems; ethical concerns regarding bias and fairness in AI applications. \\
    \hline
    Graph Prompting & 
    Seamless LLM integration; enhances performance with Knowledge Graph (KG) encoding; significant improvements in knowledge-driven tasks. & 
    Depends on quality KGs; potential noise from entity linking; requires more computational resources; variable gains. & 
    Wide application range; benefits knowledge-intensive domains; potential for complex structure handling. & 
    Risk of negative transfer; prompt design complexity; domain-specific performance variability. \\
    \hline
    Iterative Prompting & 
    Efficient LLM integration for dynamic planning. Enhances model's environmental awareness. & 
    Depends on quality data and complex environments. Introduces potential for planning noise. Requires significant computational resources. & 
    Expands applicability in real-world navigation tasks. Improves model interpretability with step-by-step reasoning. Offers potential in cross-modal applications. & 
    Complexity may hinder implementation and optimization. Performance might vary across different environments. High resource demands may limit wider use. \\
    \hline
    Least-To-Most Prompting & 
    Structured, incremental learning, versatile. & 
    Complex, not always suitable, more overhead. & 
    Wide applicability, systematic solving. & 
    Outpaced by simpler solutions, adoption barrier. \\
    \hline
    Multimodal CoT Prompting &
    Enhances multimodal reasoning by integrating visual and textual data through Chain-of-Thought (CoT) prompting. Improves performance across various benchmarks, including ScienceQA. &
    Efficiency and accuracy in reasoning across modalities, reducing hallucinations. Demonstrates scalability and generalizability across different models and tasks. &
    Complexity in integrating visual and textual reasoning may hinder broader applications beyond specific benchmarks. &
    Educational applications in developing AI tutors for complex subjects and enhancing assistive technologies for interpreting multimodal information.\\
    \hline
    ReAct Prompting &
    Integrates reasoning and action in LLMs, enhancing performance and interpretability. &
    Improves decision-making by linking reasoning to task-specific actions. &
    Outperforms on question answering and decision-making tasks, reducing errors. &
    Debate on whether improvements stem from inherent reasoning abilities or example-task similarity.\\
    \hline
    Self-Ask Prompting & 
    Versatile, improves comprehension, less data reliant. & 
    Coherence issue, complex, resource heavy. & 
    Enhances problem-solving, clear answers. & 
    Dependent on training data, prompt design is crucial. \\
    \hline
    Self-Consistency & 
    Improves performance and reliability; Reduces variability. & 
    High computational needs; Might ignore novel responses. & 
    Refines consistent response selection. & 
    Does not address underlying biases. \\
    \hline
    Sequential Prompting & 
    Precision in ranking based on user history. Leverages LLMs for nuanced recommendations. Dynamic item ranking adjustment. & 
    High computational resources. Complex to implement and maintain. Depends on initial candidate quality. & 
    Personalized user experiences. Applicable across various domains. & 
    Scalability issues due to resource demands. Data privacy and security concerns. \\
    \hline
    Tree of Thoughts (ToT) & 
    Explores multiple reasoning paths; Increases problem-solving depth. & 
    Design complexity; Identifying relevant paths challenging. & 
    Structured approach to nuanced problems. & 
    High design complexity; Information overload risk. \\
    \hline
    Zero-Shot Prompting & Perform tasks without specific examples, showcasing robust zero-shot capabilities & often requires fine-tuning or more tailored prompts & Advances in instruction tuning offer potential to enhance zero-shot performance, aligning models with human preferences. & Alternatives like few-shot learning may outperform zero-shot approaches. \\
    \hline
    \end{tabularx}
\end{table*}

\subsubsection{Tree of Thoughts (ToT)}

Tree of Thoughts (ToT) \cite{yao2023treethoughtsdeliberateproblem, long2023largelanguagemodelguided, tree-of-thought-prompting, sun2023query, sun2023reinforcement,Sahoo2024, Chen2023} is an advanced framework for enhancing language models' performance on complex tasks introduced by \cite{yao2023treethoughtsdeliberateproblem}. ToT extends chain-of-thought prompting by maintaining a hierarchical structure of intermediate steps, or "thoughts," toward solving a problem. This framework allows language models to self-evaluate progress and systematically explore different pathways using search algorithms like breadth-first search and depth-first search, incorporating lookahead and backtracking techniques. By doing so, ToT enables more effective exploration and strategic planning, improving the models' reasoning and problem-solving capabilities. Long et. al. \cite{long2023largelanguagemodelguided} built on this idea with a ToT framework that uses a reinforcement learning-trained "ToT Controller" to adapt and learn from new data, offering a more dynamic approach than traditional search methods.

Hulbert \cite{tree-of-thought-prompting} simplified the ToT concept into a single prompt technique, where language models evaluate intermediate thoughts step-by-step, making it more accessible and straightforward. Sun \cite{sun2023query, sun2023reinforcement} took ToT further with large-scale experiments and introduced PanelGPT, a creative approach that simulates panel discussions among language models to benchmark and enhance the prompting technique.

The Tree of Thoughts (ToT) framework enhances problem-solving by leveraging a tree-like search strategy. Each node in the tree represents a partial solution \( s = [x, z_{1:i}] \), where \( x \) is the initial input and \( z_{1:i} \) is the sequence of thoughts so far.

The process involves four main components:

\begin{enumerate}
    \item \textit{Thought Decomposition}: Break down the problem into manageable thought steps.
    \item \textit{Thought Generation}: Generate multiple candidates for the next thought step using strategies like i.i.d sampling or propose prompt sampling.
    \item \textit{State Evaluation}: Evaluate each state to assess progress towards the solution using heuristics or deliberate reasoning prompts.
    \item \textit{Search Algorithm}: Use search strategies like breadth-first search (BFS) or depth-first search (DFS) to explore and expand the most promising thought paths.
\end{enumerate}

Mathematically, ToT can be represented as:

\begin{equation}
s = [x, z_{1:i}]
\end{equation}

\textit{Thought Generation:}

\begin{equation}
G(p_\theta, s, k)
\end{equation}

\textit{State Evaluation:}

\begin{equation}
V(p_\theta, S)
\end{equation}

This  allows systematic exploration, lookahead, and backtracking, leveraging pre-trained language models without additional training.

\subsubsection{Zero-Shot Prompting}

Contemporary large language models (LLMs) like GPT-3.5 Turbo, GPT-4, and Claude 3 are trained on extensive datasets and optimized to follow instructions. This comprehensive training equips these models to execute tasks in a "zero-shot" fashion. Zero-shot prompting \cite{Wei2021FinetunedLM} involves giving the model a task instruction without providing any specific examples or demonstrations. The model is directly instructed to perform the task based solely on the given prompt.

\section{Metrics for Evaluating Prompt Engineering}

Prompt engineering involves the strategic formulation of inputs to guide language models (LLMs) towards desired outputs. Evaluating the efficacy of prompt engineering requires considering several key metrics.
 Table \ref{table:prompt-engineering-metrics} presents a compilation of key metrics used in the evaluation of prompt engineering. These metrics span categories such as Semantic Similarity, Diversity, and Language Acceptableness. For instance, metrics like BERTScore and STS-B focus on semantic similarity between generated and reference texts, while ROUGE and BLEU measure diversity through comparisons of n-grams and word sequences. Metrics such as CoLA and Perplexity evaluate language acceptableness and predictive performance, respectively. Understanding and applying these metrics are crucial for optimizing prompt design and enhancing the capabilities of language models in various NLP tasks.

\begin{table*}[ht!]
    \caption{Metrics for Evaluating Prompt Engineering and Corresponding Techniques}
    \label{table:prompt-engineering-metrics}
    \centering
    \renewcommand{\arraystretch}{1.3}
    \begin{tabularx}{\textwidth}{|>{\raggedright\arraybackslash}p{0.15\textwidth}|>{\raggedright\arraybackslash}p{0.35\textwidth}|>{\raggedright\arraybackslash}p{0.15\textwidth}|X|}
    \hline
    \textbf{Metric} & \textbf{Definition} & \textbf{Category} & \textbf{Corresponding Techniques} \\
    \hline
    Accuracy & Measures the proportion of correct predictions out of total predictions made by a model. & Classification & Chain-of-Thought (CoT); Few-Shot Prompting; Graph Prompting; Self-Consistency; Zero-Shot Prompting; ReAct Prompting \\
    \hline
    AUC & A metric used to evaluate the performance of a binary classification model. Represents the area under the ROC curve, plotting true positive rate against false positive rate at various thresholds. & Classification & Few-Shot Prompting, Graph Prompting; Zero-Shot Prompting \\
    \hline
    BLEU & Evaluates machine translation quality by comparing n-grams of the candidate translation with those of the reference translation. & Diversity & Few-Shot Prompting, Zero-Shot Prompting, Chain-of-Thought (CoT), Multimodal CoT Prompting \\
    \hline
    BERTScore & Uses BERT embeddings to compare the similarity between predicted and reference texts. & Semantic Similarity & Few-Shot Prompting, Zero-Shot Prompting, Chain-of-Thought (CoT) \\
    \hline
    CoLA & Evaluates language models based on their ability to determine if sentences are grammatically acceptable. & Language Acceptableness & Zero-Shot Prompting, Few-Shot Prompting \\
    \hline
    Correlation & Measures the correlation between model predictions and human annotations. & Correlation Metric & Chain-of-Thought (CoT) \\
    \hline
    F1 score & Measures the balance between precision and recall. & Classification & Few-Shot Prompting, Zero-Shot Prompting, Chain-of-Thought (CoT) \\
    \hline
    GLEU & An alternative to BLEU, considering both precision and recall of n-grams. & Diversity & Few-Shot Prompting, Zero-Shot Prompting \\
    \hline
    HIT@N (Top-N Hit Rate) & Measures how often the correct answer appears in the top-N results suggested by prompts. & Information Retrieval & Iterative Prompting, Sequential Prompting \\
    \hline
    Mean Absolute Error & Measures the average magnitude of errors between predicted and observed values. & Regression Metric & Chain-of-Thought (CoT) \\
    \hline
    METEOR & Evaluates based on precision, recall, and a synonymy matching score. & Semantic Similarity & Few-Shot Prompting, Zero-Shot Prompting, Multimodal CoT Prompting \\
    \hline
    NDCG@N (Normalized Discounted Cumulative Gain) & Evaluates the quality of ranked lists suggested by prompts, considering relevance and positions. & Information Retrieval & Graph Prompting \\
    \hline
    Perplexity & Measures how well a probability model predicts a sample, with lower perplexity indicating better performance. & Language Acceptableness & Few-Shot Prompting, Zero-Shot Prompting, Self-Consistency \\
    \hline
    Recall & Measures the proportion of actual positives correctly identified by the model. & Classification & Graph Prompting \\
    \hline
    ROUGE & Evaluates automatic summarization by comparing overlapping units between the generated summary and the reference summary. & Diversity & Few-Shot Prompting, Zero-Shot Prompting, Chain-of-Thought (CoT), Tree of Thoughts (ToT) \\
    \hline
    STS-B & Measures the degree of semantic equivalence between sentence pairs. & Semantic Similarity & Chain-of-Thought (CoT), Few-Shot Prompting \\
    \hline
    \end{tabularx}
\end{table*}

\section{Conclusion}

In conclusion, this survey paper has provided a comprehensive analysis of prompt engineering techniques within the context of Large Language Models (LLMs). By conducting a SWOT analysis, we have highlighted the strengths, weaknesses, opportunities, and threats associated with various methods such as zero-shot prompting, few-shot prompting, chain-of-thought prompting, and more. Our findings underscore the critical role of linguistic principles in shaping effective prompt design and the potential of these techniques to enhance AI interactions and understanding of human prompts. The key findings include identifying synergies between AI, Linguistics, and Prompt Engineering, categorizing numerous prompt engineering methods, identifying metrics such as BLEU, BERTScore, ROUGE, and Perplexity for evaluation, and conducting a SWOT analysis of various prompt engineering techniques. Despite the notable advancements, challenges such as prompt complexity, computational demands, and domain-specific limitations persist. Future research should focus on addressing these challenges, optimizing prompt engineering strategies, and exploring novel applications to further improve the efficacy and reliability of LLMs in diverse real-world scenarios.





\bibliographystyle{IEEEtran}
\bibliography{ref.bib}

\end{document}